\documentclass[pdflatex,sn-mathphys-num]{sn-jnl}

\usepackage{graphicx}%
\usepackage{multirow}%
\usepackage{amsmath,amssymb,amsfonts}%
\usepackage{amsthm}%
\usepackage{mathrsfs}%
\usepackage[title]{appendix}%
\usepackage{xcolor}%
\usepackage{textcomp}%
\usepackage{manyfoot}%
\usepackage{booktabs}%
\usepackage{algorithm}%
\usepackage{algorithmicx}%
\usepackage{algpseudocode}%
\usepackage{listings}%

\theoremstyle{thmstyleone}%
\newtheorem{theorem}{Theorem}
\theoremstyle{thmstyletwo}%

\theoremstyle{thmstylethree}%
\newtheorem{corollary}{Corollary}[theorem]

\begin{document}

\title[Online Neural Networks for Change-Point Detection]{Online Neural Networks for Change-Point Detection}


\author*[1]{\fnm{Mikhail} \sur{Hushchyn}}\email{mhushchyn@hse.ru}

\author[1]{\fnm{Kenenbek} \sur{Arzymatov}}\email{kenenbek@gmail.com}

\author[1]{\fnm{Denis} \sur{Derkach}}\email{dderkach@hse.ru}

\affil[1]{\orgname{HSE University}, \orgaddress{\city{Moscow}, \country{Russia}}}




\abstract{Moments when a time series changes its behavior are called change points. Occurrence of change point implies that the state of the system is altered and its timely detection might help to prevent unwanted consequences. In this paper, we present two change-point detection approaches based on neural networks and online learning. These algorithms demonstrate linear computational complexity and are suitable for change-point detection in large time series. We compare them with the best known algorithms on various synthetic and real world data sets. Experiments show that the proposed methods outperform known approaches. We also prove the convergence of the algorithms to the optimal solutions and describe conditions rendering current approach more powerful than offline one.}




\maketitle

\section{Introduction}

The first works~\cite{10.2307/2333009, 10.1093/biomet/42.3-4.523} about change-point detection were presented in the 1950s. They utilize shifts of the signal mean value to detect changes in the quality of the output of a continuous production process. In the following decades, a lot of other change-point detection methods were developed. They are based on different ideas and are able to recognize various changes in time series: jumps of mean and variance of a signal, correlations between its different components and other more elaborate dependencies. These algorithms are well-described in various overviews~\cite{10.5555/151741, Aminikhanghahi2017, TRUONG2020107299}. Detection of change points can be found in many applications: quality monitoring of industrial processes, failure detection in complex systems, health monitoring, speech recognition and video analysis.

This study introduces two new approaches for change-point detection based on online learning of neural networks. These algorithms can be used to detect changes in time series behavior. As it is shown in the following sections, they have linear computational complexity, work with multidimensional signals and are well suited for long time series. The proposed solutions are inspired by the Kullback–Leibler importance estimation procedure (KLIEP)~\cite{kliep1}, unconstrained least-squares importance fitting (uLSIF)~\cite{ulsif1, Sugiyama2011} and the relative uLSIF (RuLSIF)~\cite{rulsif1, Yamada2013}. The methods are used to estimate the direct probability density ratio for two samples. As demonstrated in~\cite{LIU201372}, this approach can be used for change-point detection in time series data. Moreover, according to~\cite{Aminikhanghahi2017}, better results are expected with respect to other change-point detection algorithms. The idea is based on calculation of distances between pairs of observations from two different samples using Radial basis function (RBF) kernels to approximate the probability density ratio. 

The first implementation of decision tree and logistic regression classifiers to analyze changes between two samples is demonstrated in~\cite{Hido2008}. However, the method is not applied for change-point detection. The authors of~\cite{Nam2015} show that Convolutional  Neural  Networks (CNNs), trained with uLSIF loss function can be used for outlier detection in images. In recent years, several approaches based on neural networks~\cite{khan2019deep, hushchyn2020generalization}, with KLIEP and RuLSIF loss functions, were presented for change-point detection in time series data. It is also shown that they outperform previous methods based on RBF kernels.

\section{Change-Point Detection}

\begin{figure}
\centering
\includegraphics[width=1.\linewidth]{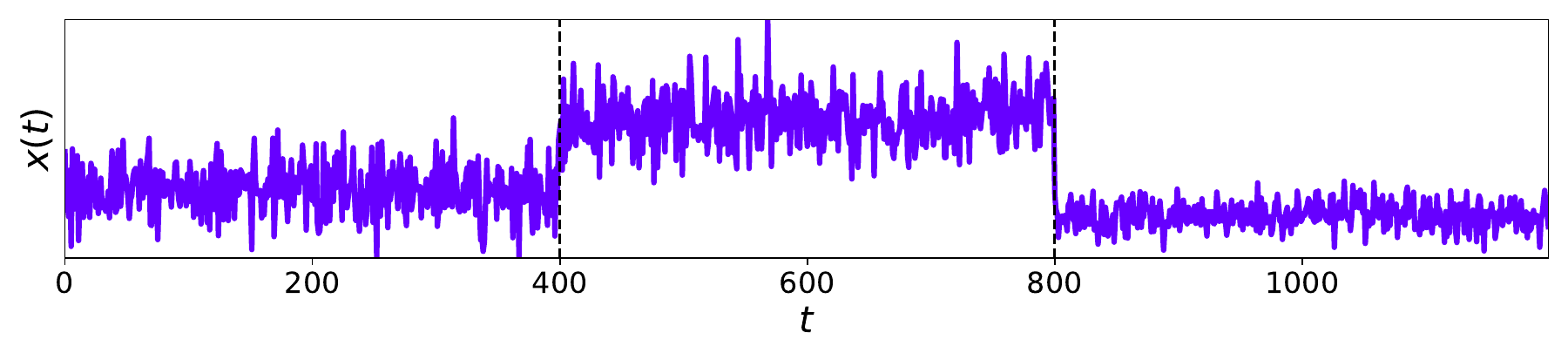}
\caption{Example of a time series with two change-points at moments $t_{1}=400$ and $t_{2}=800$. Observations between these points have different probability distributions: $P_{1}(x(t))$ for $0 < t < t_{1}$, $P_{2}(x(t))$ for $t_{1} < t < t_{2}$ and $P_{3}(x(t))$ for $t_{2} < t < 1200$.}
\label{fig:example}
\end{figure}

Consider a time series, where each observation for a moment $t$ is represented by a $d-$dimensional vector $x(t) \in \mathcal{R}^{d}$:

\begin{equation}
    x(1), x(2), x(3), ..., x(\tau), x(\tau+1), x(\tau+2), ...
\label{eq:ts1}
\end{equation}

Assume that all observations $x(t)$ with $t < \tau$ have probability density distribution $p_0(x)$, and all observations with $t \ge \tau$ are sampled from distribution $p_1(x) \ne p_0(x)$. In other words, the time series changes its behavior at moment $\tau$. Such moments are called change-points. There may be several such points in one time series, as it is demonstrated in Figure~\ref{fig:example}. The goal is to detect all change points with the highest quality. This is an unsupervised problem, since the true positions of change-points are not given.

Often the original time series is transformed into an autoregression form~\cite{LIU201372}:

\begin{equation}
    X(k), X(k+1), X(k+2), ..., X(\tau), X(\tau+1), X(\tau+2), ...
\label{eq:ts2}
\end{equation}

where $X(t)$ is a combined vector of $k$ previous observations of the time series and is defined as:

\begin{equation}
    X(t) = [x(t)^{T}, x(t-1)^{T}, ..., x(t-k+1)^{T}]^{T} \in \mathcal{R}^{kd}
\label{eq:ar}
\end{equation}

This transformation allows us to take into account time dependencies between observations and helps to improve the quality of change-point detection. It is equal to the time series in Eq.~\ref{eq:ts1} with $k=1$. We also use this notation to preserve consistency with conventional notation.

\section{Quality Metrics}

Consider a time series with $n$ change-points at moments $\tau_{1}$, $\tau_{2}$, ..., $\tau_{n}$. Suppose that an algorithm recognises $m$ change-points at moments $\hat{\tau}_{1}$, $\hat{\tau}_{2}$, ..., $\hat{\tau}_{m}$. Following~\cite{TRUONG2020107299}, a set of correctly detected change-points is defined as True Positive (TP):

\begin{equation}
    \text{TP} = \{ \tau_{i} | \exists \hat{\tau}_{j}: |\hat{\tau}_{j} - \tau_{i}| < M \}
\end{equation}

where $M$ is a margin size and $M=50$ in our study. Then, Precision, Recall and F1-score metrics are calculated as follows:

\begin{equation}
    \text{Precision} = \frac{|\text{TP}|}{m}
\end{equation}

\begin{equation}
    \text{Recall} = \frac{|\text{TP}|}{n}
\end{equation}

\begin{equation}
    \text{F1} = \frac{2 \cdot \text{Precision} \cdot \text{Recall}}{\text{Precision} + \text{Recall}}
\end{equation}

We use F1-score to measure quality of change-point detection algorithms. We also use a common measure in clustering analysis, called Rand Index (RI)~\cite{rand_index}, which is calculated in the following way. True change-points $\{ \tau_i \}_n$ split the time series into $n+1$ segments $S$. Similarly, the observations are divided by the detected change-points $\{ \hat{\tau}_i \}_m$ into $m+1$ segments $\hat{S}$. RI measures the similarity of these two segmentation sets. The Rand Index is then defined as

\begin{equation}
    \text{RI} = \frac{A}{0.5~ T(T-1)},
\end{equation}

where $A$ is the number of observation pairs $x(i)$ and $x(j)$, that share the same segment, both in  $S$ and $\hat{S}$; $T$ is the total number of observations in the time series and $0.5~T (T-1)$ gives the total number of observation pairs in the whole time series.

\section{Proposed Methods}

\begin{figure}
\centering
\includegraphics[width=1.\linewidth]{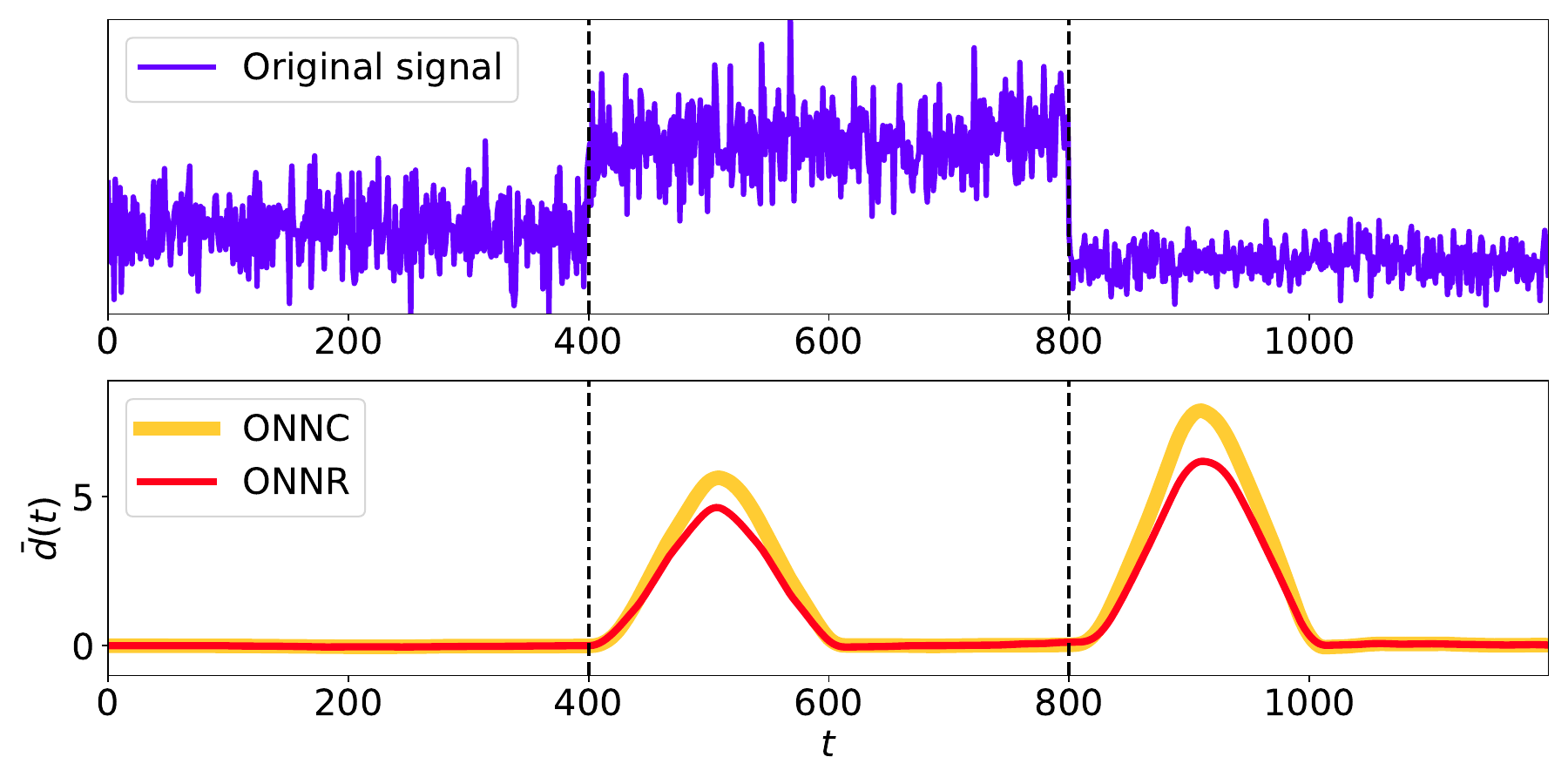}
\caption{Example of change-point detection using the proposed algorithms. (Top) A time series with two change-points at moments $t_{1}=400$ and $t_{2}=800$. (Bottom) Change-point detection score $\bar{d}(t)$ estimated by the algorithms ONNC and ONNR.}
\label{fig:cpd_example}
\end{figure}

\subsection{Classification-Based Model}

Consider a time series defined in Eq.~\eqref{eq:ts2} with several change-points. The idea of the proposed algorithm is based on a comparison of two observations $X(t-l)$ and $X(t)$ of this time series. Here $l$ is the lag size between these two observations. If there is no change point between them, $X(t-l)$ and $X(t)$ have the same distributions. Otherwise, they are sampled from different distributions, which means that a change point occurred at the moment $\tau: t-l < \tau \le t$. Repeating this comparison for all pairs of observations sequentially helps to determine the positions of all change-points in the time series.  

A more general way is to compare two mini-batches of observations $\mathcal{X}(t-l)$ and $\mathcal{X}(t)$. Here, a mini-batch $\mathcal{X}(t)$, is a sequence of observations of size $n$, which is defined as:

\begin{equation}
\mathcal{X}(t) = \{ X(t), X(t-1), ..., X(t-n+1) \}
\end{equation}

Further in this study, we work with these mini-batches of size $n \ll l$ in order to speed up the change-point detection algorithm. 

To check whether observations in two mini-batches $\mathcal{X}(t-l)$ and $\mathcal{X}(t)$ come from the same distribution, we use a classification model based on a neural network $f(X, \theta)$ with weights $\theta$. This network is trained on the mini-batches with cross-entropy loss function $L_t(\theta)$,

\begin{equation}
L_t(\theta) = - \frac{1}{n} \sum_{X \in \mathcal{X}(t-l)} \log (1 - f(X, \theta)) -  \frac{1}{n} \sum_{X \in \mathcal{X}(t)} \log f(X, \theta),
\end{equation}
where all observations from $\mathcal{X}(t-l)$ are considered as the negative class and observations from $\mathcal{X}(t)$ are taken as the positive class. We use only one neural network for the whole time series and it is trained in accordance with the online learning paradigm: each pair of mini-batches is used only once and the network makes a few iterations of optimization on each pair. Information from previous pairs are encoded in the neural network weights and each new step just slightly changes them. 

The neural network $f(X, \theta)$ can be used to compare distributions of observations in the mini-batches. In this work, we use a dissimilarity score based on the  Kullback-Leibler divergence, $D_t(\theta)$. Following~\cite{hushchyn2020generalization}, we define this score as

\begin{equation} 
\label{eq:d_clf}
\begin{split}
    D_t(\theta) = & \frac{1}{n} \sum_{X \in \mathcal{X}(t-l)} \log \frac{1 - f(X, \theta)}{f(X, \theta)} + \\
     + & \frac{1}{n} \sum_{X \in \mathcal{X}(t)} \log \frac{f(X, \theta)}{1 - f(X, \theta)}.
\end{split}
\end{equation}

If observations in the mini-batches are sampled from the same distribution, this dissimilarity score value is close to 0. Otherwise, it takes positive values. All steps above are combined into one algorithm called change-point detection based on Online Neural Network Classification (ONNC) and shown in Algorithm~\ref{alg:clf}. An example of change-point detection, using ONNC, is demonstrated in Figure~\ref{fig:cpd_example}.

\begin{algorithm}[t!]
\caption{ONNC change-point detection algorithm.}
\begin{algorithmic}[1]
\State \textbf{Inputs:} time series $\{X(t)\}_{t=k}^{T}$; $k$ -- size of a combined vector $X(t)$; $n$ -- size of a mini-batch $\mathcal{X}(t)$; $l$ -- lag size and $n \ll l$; $f(X, \theta)$ -- a neural network with weights $\theta$;
\State \textbf{Initialization:} $t \leftarrow k + n + l$;
\While{$t \le T$}
 \State take mini-batches $\mathcal{X}(t-l)$ and $\mathcal{X}(t)$;
 \State $d(t) \leftarrow D_t(\theta)$;
 \State $\bar{d}(t) \leftarrow \bar{d}(t-n) + \frac{1}{l} (d(t) - d(t-l-n))$;
 \State $\mathrm{loss}(t, \theta) \leftarrow L_t(\theta)$;
 \State $\theta \leftarrow \mathrm{Optimizer}(\mathrm{loss}(t, \theta))$;
 \State $t \leftarrow t + n$;
\EndWhile
\State \Return{$\{\bar{d}(t)\}_{t=1}^{T}$} -- change-point detection score
\end{algorithmic}
\label{alg:clf}
\end{algorithm}

\subsection{Regression-Based Model}

An alternative method of change-point detection is based on regression models. In this case, a regression model, based on a neural network $g(X, \theta)$, with weights $\theta$, is used to estimate the ratio between distributions of a time series observations in two mini-batches $\mathcal{X}(t-l)$ and $\mathcal{X}(t)$. Assume that all observations in $\mathcal{X}(t-l)$ have a probability density distribution $q(X)$, and observations in $\mathcal{X}(t)$ mini-batch are sampled from the distribution $p(X)$. Then, the output of the neural network approximates the ratio between these two distributions directly

\begin{equation}
g(X, \theta) \approx \frac{p(X)}{q(X)}.
\end{equation}

Following the idea of the RuLSIF method~\cite{rulsif1, Yamada2013} and mathematical inference in~\cite{hushchyn2020generalization}, the loss function for the neural network is defined as

\begin{equation}
\label{eq:j_rulsif}
\begin{split}
L(\mathcal{X}(t-l), \mathcal{X}(t), \theta) & = \frac{1-\alpha}{2n} \sum_{X \in \mathcal{X}(t-l)} g^{2}(X, \theta) + \\
& + \frac{\alpha}{2n} \sum_{X \in \mathcal{X}(t)} g^{2}(X, \theta) - \frac{1}{n} \sum_{X \in \mathcal{X}(t)} g(X, \theta),
\end{split}
\end{equation}

where $\alpha$ is an adjustable parameter. In this work, we take $\alpha=0.1$. Similarly to the classification-based algorithm, described in the previous section, the neural network is trained in an online learning way: all mini-batches are processed only once in time order.

While the output $g(X, \theta)$ approximates the ratio between the distributions of observations in the mini-batches, we can estimate the dissimilarity score between them using the Pearson $\chi^{2}-$divergence~\cite{hushchyn2020generalization}:

\begin{equation} \label{eq:d_reg}
\begin{split}
D(\mathcal{X}(t-l), \mathcal{X}(t), \theta) = \frac{1}{n} \sum_{X \in \mathcal{X}(t)} g(X, \theta) - 1
\end{split}
\end{equation}

However, the loss function and the dissimilarity score described above are asymmetric with respect to the mini-batches $\mathcal{X}(t-l)$ and $\mathcal{X}(t)$, and affect the change-point detection quality. To compensate this effect, we use two neural networks $g_1(X, \theta_1)$ and $g_2(X, \theta_2)$ as is described in Algorithm~\ref{alg:reg}. We call this algorithm change-point detection based on Online Neural Network Regression (ONNR). An example of change-point detection using this algorithm is shown in Figure~\ref{fig:cpd_example}.

\begin{algorithm}[t!]
\caption{ONNR change-point detection algorithm.}
\begin{algorithmic}[1]
\State \textbf{Inputs:} time series $\{X(t)\}_{t=k}^{T}$; $k$ -- size of a combined vector $X(t)$; $n$ -- size of a mini-batch $\mathcal{X}(t)$; $l$ -- lag size and $n \ll l$; $g_1(X, \theta_1)$ and $g_2(X, \theta_2)$ -- neural network with weights $\theta_1$ and $\theta_2$ respectively;
\State \textbf{Initialization:} $t \leftarrow k + n + l$;
\While{$t \le T$}
 \State take mini-batches $\mathcal{X}(t-l)$ and $\mathcal{X}(t)$;
 \State $d_1(t) \leftarrow D(\mathcal{X}(t-l), \mathcal{X}(t), \theta_1)$;
 \State $d_2(t) \leftarrow D(\mathcal{X}(t), \mathcal{X}(t-l), , \theta_2)$;
 \State $d(t) \leftarrow d_1(t) + d_2(t)$;
 \State $\bar{d}(t) \leftarrow \bar{d}(t-n) + \frac{1}{l} (d(t) - d(t-l-n))$;
 \State $loss(t, \theta_1) \leftarrow L(\mathcal{X}(t-l), \mathcal{X}(t), \theta_1)$;
 \State $\theta_1 \leftarrow \mathrm{Optimizer}_1(\mathrm{loss}(t, \theta_1))$;
 \State $loss(t, \theta_2) \leftarrow L(\mathcal{X}(t), \mathcal{X}(t-l), \theta_2)$;
 \State $\theta_2 \leftarrow \mathrm{Optimizer}_2(\mathrm{loss}(t, \theta_2))$;
 \State $t \leftarrow t + n$;
\EndWhile
\State \Return{$\{\bar{d}(t)\}_{t=1}^{T}$} -- change-point detection score
\end{algorithmic}
\label{alg:reg}
\end{algorithm}

\section{Properties of the Algorithm}
\label{sec:theory}

\subsection{Convergence Properties}

In this section, we describe several theoretical properties of the ONNC algorithm in a special case. We consider the batch size of $n=1$ and $X(i) = x(i)$ for simplicity. As a result we fit the algorithm with cross-entropy loss function $L_t(\theta)$ in the following form:

\begin{equation}
L_t(\theta) = - \log (1 - f(x(t-l), \theta)) -  \log f(x(t), \theta).
\end{equation}

We also consider the last $N$ steps of the algorithm and the lag size $l=N$ for further analysis and change-point detection score estimation:

\begin{equation}
I_N = \sum_{i=t-N+1}^{t} L_{i}(\theta_i).
\end{equation}

Without losing the generality, we suppose that we use Online Gradient Descent (OGD) algorithm for the optimization:
\begin{equation}
\theta_i = \theta_{i-1} - \eta \nabla_{\theta} L_{i}(\theta_{i-1})
\end{equation}
where $\eta$ is a learning rate. We also assume that $\theta \in F$. We follow the assumptions for the feasible set $F$ and $L_i(\theta_i)$ functions as described in~\cite{10.5555/3041838.3041955}. To explore the algorithm's properties, we analyze the regret of this algorithm for the last $N$ steps:
\begin{equation}
R(N) = \sum_{i=t-N+1}^{t} L_{i}(\theta_i) - \min_{\theta} \sum_{i=t-N+1}^{t} L_{i}(\theta),
\end{equation}
where the first term corresponds to the OGD algorithm. The second term corresponds to the offline algorithm~\cite{hushchyn2020generalization}, that finds a static feasible solution for the all $N$ steps. Regard the following theorem for the offline optimization, and the theorem, that shows when the online algorithm outperform the offline one.

\begin{theorem}
\label{lem:1}
For any time moment $t-N < \nu \le t$ the following inequality holds:
\begin{equation}
\min_{\theta} \sum_{i=t-N+1}^{t} L_{i}(\theta) \ge \min_{\theta} \sum_{i=t-N+1}^{\nu} L_{i}(\theta) + \min_{\theta} \sum_{i=\nu+1}^{t} L_{i}(\theta)
\end{equation}
\end{theorem}

\begin{theorem}
\label{theor:1}
For any time moment $t-N < \nu \le t$ the follwoing inequality holds:
\begin{equation}
R(N) \le \frac{\parallel F \parallel^2}{\eta} + \frac{\parallel \nabla L \parallel^2}{2} \eta N - C(N, \nu)
\end{equation}
where
\begin{equation}
\begin{split}
    \parallel F \parallel & = \max_{x, y \in F} d(x, y)
\end{split}
\end{equation}
\begin{equation}
\begin{split}
    \parallel \nabla L \parallel & = \max_{x \in F, i \in \{t, t-1, ...\}} \parallel \nabla L_i(\theta_{i-1}) \parallel
\end{split}
\end{equation}
\begin{equation}
\begin{split}
    C(N, \nu) & = \min_{\theta} \sum_{i=t-N+1}^{t} L_{i}(\theta)  \\
              & - \min_{\theta} \sum_{i=t-N+1}^{\nu} L_{i}(\theta) - \min_{\theta} \sum_{i=\nu+1}^{t} L_{i}(\theta) \ge 0
\end{split}
\end{equation}

\end{theorem}

This theorem defines the upper bound for the online algorithm regret. It helps to estimate conditions when the online algorithm finds smaller loss function values compared to the offline one. These conditions are presented in the following corollaries of the Theorem \ref{theor:1}. The proof of this theorem is provided in Appendix \ref{sec:proofs}.

\begin{corollary}
\label{cor:eta}
For a given $N$ and learning rate $\eta = \sqrt{\frac{2\parallel F \parallel^2}{N\parallel \nabla L(\theta) \parallel^2}}$ the upper bound of the regret $R(N)$ reaches its minimum:
\begin{equation}
R(N) \le \sqrt{2N \parallel F \parallel^2 \parallel \nabla L(\theta) \parallel^2} - C(N, \nu)
\end{equation}
\end{corollary}

\begin{corollary}
\label{cor:regret}
For a given $N$, learning rate $\eta = \sqrt{\frac{2\parallel F \parallel^2}{N\parallel \nabla L(\theta) \parallel^2}}$, and $C(N, \nu) > \sqrt{2N \parallel F \parallel^2 \parallel \nabla L(\theta) \parallel^2}$ the regret $R(N)$ takes negative values:
\begin{equation}
R(N) < 0
\end{equation}
\end{corollary}

\begin{corollary}
\label{cor:rc}
For a given $N$, learning rate $\eta = \sqrt{\frac{2\parallel F \parallel^2}{N\parallel \nabla L(\theta) \parallel^2}}$, and $\nu^{*}=\arg \max_{\nu} C(N, \nu)$ the upper bound of $R(N) + C(N, \nu^{*})$ reaches its minimum:
\begin{equation}
R(N) + C(N, \nu^{*}) \le \sqrt{2N \parallel F \parallel^2 \parallel \nabla L(\theta) \parallel^2}.
\end{equation}
Therefore, 
\begin{equation}
\lim_{N\to\infty}\frac{R(N) + C(N, \nu^{*})}{N} \le 0.
\end{equation}
\end{corollary}

The Corollary \ref{cor:eta} estimates the optimal learning rate for OGD algorithm used for change-point detection. The Corollary \ref{cor:regret} defines the conditions when OGD has a lower loss function value than the offline optimization algorithm. The Corollary \ref{cor:rc} shows estimation of convergence of the online algorithm. Intuition for these results is that online algorithms adapts to changes of signal distribution and finds lower loss function value.

\subsection{Offline and Online Dissimilarity Scores}

Now consider a general example with a change-point at the time moment $t-\nu+1$. In this case, $x(i) \sim p_0(x)$ for $i \le t-\nu$ and $x(i) \sim p_1(x)$ for $i > t+\nu$. We compare change point detection scores for the optimal offline and online change point detection algorithms. The offline algorithm corresponds to the minimum of the loss function:
\begin{equation}
I_N^{offline} = \min_{\theta} \sum_{i=t-N+1}^{t} L_{i}(\theta).
\end{equation}

The optimal online algorithm provides the minimum of the following expression as it shown in Corollary \ref{cor:rc}:
\begin{equation}
I_N^{online} = \min_{\theta} \sum_{i=t-N+1}^{t-\nu} L_{i}(\theta) + \min_{\theta} \sum_{i=t-\nu+1}^{t} L_{i}(\theta).
\end{equation}

For both methods we take the average dissimilarity score for the change point detection:
\begin{equation} 
\label{eq:d_clf_2}
\begin{split}
    \bar{d}(t) = \frac{1}{N} \sum_{t-N+1}^{t} \left( \log \frac{1 - f(x(i-N), \theta_i)}{f(x(i-N), \theta_i)} + \log \frac{f(x(i), \theta_i)}{1 - f(x(i), \theta_i)} \right).
\end{split}
\end{equation}

\begin{theorem}
\label{theor:2}
For any time moment $t$ and $v \le N$ the following equation holds for the online algorithm:
\begin{equation}
\mathbb{E}[\bar{d}(t)^{online}] = \frac{\nu}{N} \left( \mathbb{E}_{x\sim p_1(x)} \left[ \log \frac{p_1(x)}{p_0(x)} \right] - \mathbb{E}_{x\sim p_0(x)} \left[ \log \frac{p_1(x)}{p_0(x)} \right] \right).
\end{equation}
\end{theorem}

\begin{theorem}
\label{theor:3}
For any time moment $t$ and $v \le N$ the following equation holds for the offline algorithm:
\begin{equation}
\mathbb{E}[\bar{d}(t)^{offline}] = \frac{\nu}{N} \left( \mathbb{E}_{x\sim p_1(x)} \left[ \log \frac{\tilde{p}_1(x)}{\tilde{p}_0(x)} \right] - \mathbb{E}_{x\sim p_0(x)} \left[ \log \frac{\tilde{p}_1(x)}{\tilde{p}_0(x)} \right] \right).
\end{equation}
where
\begin{equation}
\frac{\tilde{p}_1(x)}{\tilde{p}_0(x)} = 1 + \frac{\nu}{N} \left(\frac{p_1(x)}{p_0(x)}-1\right).
\end{equation}
\end{theorem}

\begin{corollary}
\label{cor:off_on_1}
For $\frac{\nu}{N} \left(\frac{p_1(x)}{p_0(x)}-1\right) << 1$, $\mathbb{E}[\bar{d}(t)^{offline}] = O(\nu^2)$.
\end{corollary}

\begin{corollary}
\label{cor:off_on_2}
For any time moment $t$ and $v \le N$ the following inequality holds for the offline and online algorithms:
\begin{equation}
\mathbb{E}[\bar{d}(t)^{online}] \ge \mathbb{E}[\bar{d}(t)^{offline}].
\end{equation}
\end{corollary}

\begin{figure}
\centering
\includegraphics[width=1.\linewidth]{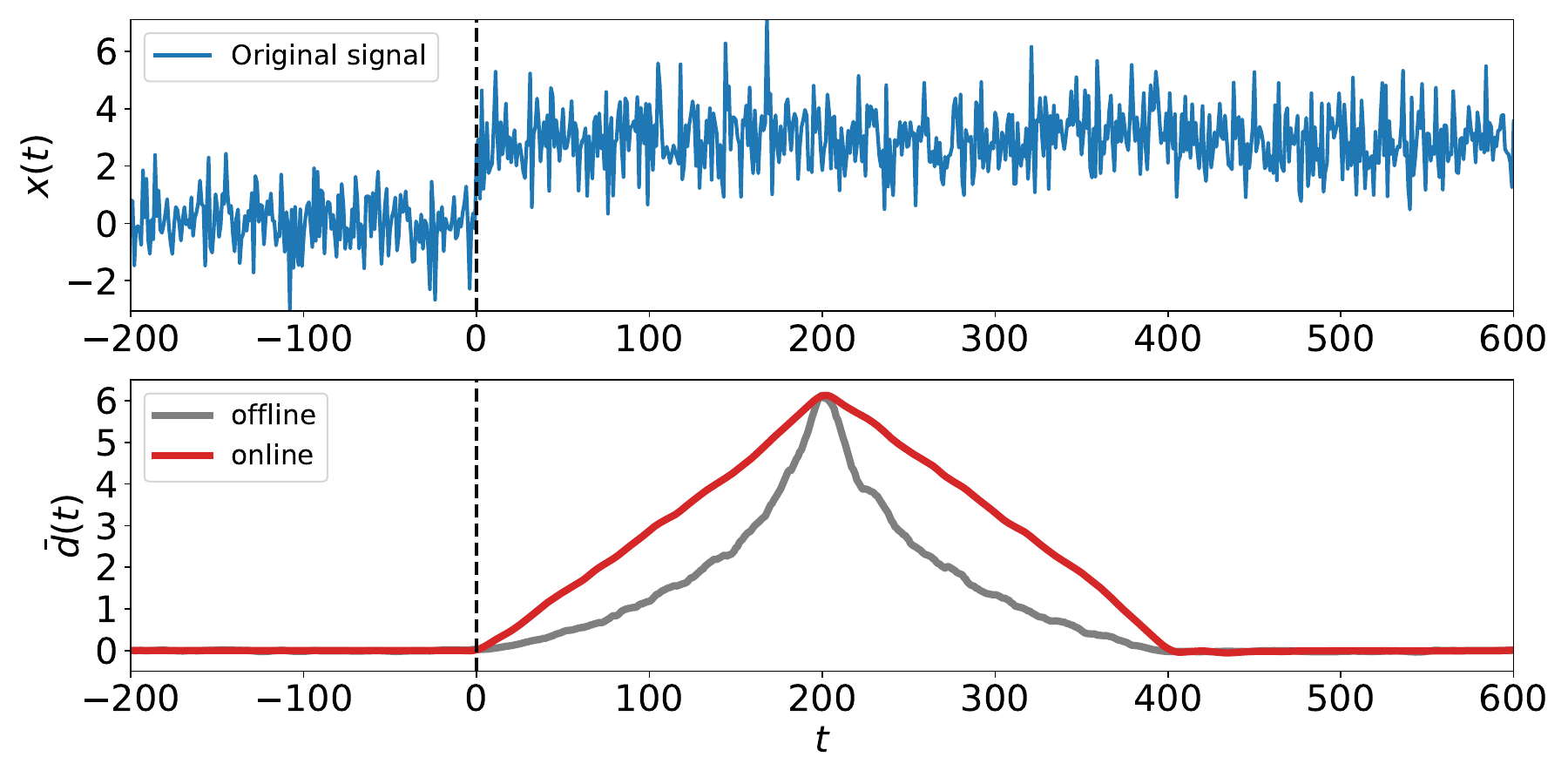}
\caption{Example of change-point detection using the online and offline algorithms. (Top) A time series with one change-point at moments $t=0$. (Bottom) Change-point detection score $\bar{d}(t)$ estimated by the algorithms.}
\label{fig:online_offline}
\end{figure}

The proof of the theorems is provided in Appendix \ref{sec:proofs}. Example of change-point detection using the online and offline algorithms is shown in Figure \ref{fig:online_offline}. In this example a change point occurs at time moment $t=0$ with $p_0(x)=\mathcal{N}(0, 1)$ for $t < 0$ and $p_1(x)=\mathcal{N}(3, 1)$ for $t \ge 0$. The figure demonstrates the relation between the dissimilarity scores $\bar{d}(t)$ for the online and offline algorithms, that is defined by the Theorems \ref{theor:2} and \ref{theor:3} for $N=200$.

\section{Data Sets}
\label{datasets}

To test change-point detection algorithms, we use several synthetic and real world data sets with various numbers of dimensions. Their purpose is to estimate how different methods work in different conditions and with different kinds of change-points. The first synthetic data set is called \textbf{mean jumps} and contains 10 one-dimensional time series, where each observation $x(t)$ is sampled from normal distribution $x(t) \sim \mathcal{N}(\mu, \sigma)$ with mean $\mu$ and standard deviation $\sigma=1$. Change-points are generated every 200 timestamps by changing mean $\mu$ in the following way:

\begin{equation}
\mu_{N} = 
\begin{cases} 
0, & \mbox{if } N = 1 \\ 
\mu_{N-1} + 0.2N, & \mbox{if } N = 2, ..., 10, 
\end{cases}
\end{equation}

where $N$ is an integer which is estimated as $200(N-1) < t \le 200N$.

Similarly, \textbf{variance jumps} data set contains 10 one-dimensional time series, where each observation $x(t)$ is also sampled from normal distribution $x(t) \sim \mathcal{N}(\mu, \sigma)$ with mean $\mu=0$ and standard deviation $\sigma$. Change-points are generated every 200 timestamps by changing $\sigma$ in the following way:

\begin{equation}
\sigma_{N} = 
\begin{cases} 
1, & \mbox{if } N = 2k+1 \\ 
1 + 0.25N, & \mbox{if } N = 2k 
\end{cases}
\end{equation}

where $N$ is an integer that is estimated as $200(N-1) < t \le 200N$.

The last synthetic data set we use in this work is called \textbf{cov jumps}. It also contains 10 two-dimensional time series, where each observation $x(t)$ is sampled from multivariate normal distribution $x(t) \sim \mathcal{N}(\mu, \Sigma)$, with a vector of means $\mu=(0, 0)^{T}$ and covariance matrix $\Sigma$. As previously, change-points are generated every 200 timestamps by changing $\Sigma$ in the following way:

\begin{equation}
\Sigma_{N} = 
\begin{cases} 
\begin{pmatrix}
1 & -0.1N \\
-0.1N & 1 \\        
\end{pmatrix}, & \mbox{if } N = 2k+1 \\ 
\begin{pmatrix}
1 & 0.1N \\
0.1N & 1 \\        
\end{pmatrix}, & \mbox{if } N = 2k 
\end{cases}
\end{equation}

where $N$ is an integer that is estimated as $200(N-1) < t \le 200N$.

We also use two real world data sets that are publicly available and are taken from the human activity recognition domain. \textbf{WISDM}~\cite{8835065, Dua:2019} data set contains 3-dimensional signals of accelerometer and gyroscope sensors, collected from a smartphone and a smartwatch measured at a rate of 20 Hz. The signal is collected for different human activities. Their changes are considered as change-points. Each time series has 17 change-points. We use 10 samples of the smartwatch gyroscope sensors for further tests. We also downsample the signals and take only about 3000 observations per time series.

Similarly, \textbf{EMG} Physical Action Data Set~\cite{Dua:2019} contains EMG data, which corresponds to 10 different physical activities for 4 persons. Transitions between the activities are considered as change-points. Each sample has 8 dimensions. We downsample the original signals to only about 2000 measurements per time series for the change-point detection tests.

One more interesting data set we use is called \textbf{Kepler}~\cite{kepler2019}. It contains data from the Kepler spacecraft that was launched in March 2009. Its mission was to search for transit-driven exoplanets, located within the habitable zones of Sun-like stars. In this work we use the one-dimensional Kepler light curves, with Data Conditioning Simple Aperture Photometry (DCSAP) data from 10 stars with exoplanets.

The next range of data sets are based on real samples for classification tasks in machine learning, collected from astronomical and high energy physics domains. 

The first data set is called \textbf{HTRU2}~\cite{10.1093/mnras/stw656, Dua:2019} and describes a sample of pulsar candidates, collected during the High Time Resolution Universe Survey (South)~\cite{doi:10.1111/j.1365-2966.2010.17325.x}. It contains two types of astronomical objects: positive (pulsars) and negative (others), that are described by 8 features. We create 10 time series with 2000 observations $x(t)$, that are sampled from positive or negative classes with change-points at every 200 timestamps:

\begin{equation}
\label{eq:clf2ser}
x(t) = 
\begin{cases} 
\text{random negative object}, & \mbox{if } N = 2k  \\
\text{random positive object}, & \mbox{if } N = 2k+1 
\end{cases}
\end{equation}

where $N$ is an integer that is estimated as $200(N-1) < t \le 200N$. Changes of the object classes are considered as change-points. Then, we scale each components of the time series by reducing their mean values to 0 and variance to 1. After that, we add white noise generated from the normal distribution $\mathcal{N}(\mu=0, \sigma=2)$. The goal of this transformation is to reduce the difference between the distributions of the classes and make change-point detection more difficult.

One more astronomical data set is \textbf{MAGIC} Gamma Telescope Data Set~\cite{Dua:2019}, which describes signals registered in the Cherenkov gamma telescope, from high energy particles, that come from space. There are also two kinds of signals: positive and negative, that correspond to gamma and hadron particles respectively. Each 
signal is described by 10 features. Similar to the HTRU2 data set, we create 10 time series by sampling observations $x(t)$ as is shown in~\eqref{eq:clf2ser} and adding noise generated from $\mathcal{N}(\mu=0, \sigma=5)$ to each component.

\textbf{SUSY}~\cite{Baldi_2014, Dua:2019} is a data set from a high energy physics domain. It contains positive (signal) and negative (background) events, observed in a particle detector and described by 18 features. We create 10 time series in the same way as for the HTRU2 data set.

One more high energy physics data set is called \textbf{Higgs}~\cite{Baldi_2014, Dua:2019} and contains positive (signal) and negative (background) events. Each event is described by 21 features. While it is a quite difficult data set for change-point detection, we create 10 time series with 4000 observations $x(t)$, that are sampled from the positive or negative classes:

\begin{equation}
x(t) = 
\begin{cases} 
\text{random negative object}, & \mbox{if } N = 2k  \\
\text{random positive object}, & \mbox{if } N = 2k+1 
\end{cases}
\end{equation}

where $N$ is an integer that is estimated as $400(N-1) < t \le 400N$.  Changes of the object classes are considered as change-points.

The final data set we use in this work is \textbf{MNIST}~\cite{Dua:2019}, which contains 1794 samples of hand-written digits. Each digit is described by 64 features. We create 10 time series with 1794 observations $x(t)$ by stacking all randomly shuffled 0 digits, then adding all randomly shuffled 1 digits and repeating this for all classes. Changes of the digits are considered as change-points. Then, similarly to the HTRU2 data set, we add white noise, generated from normal distribution $\mathcal{N}(\mu=0, \sigma=5)$.

\section{Experiments}

\begin{figure}[t]
\centering
\includegraphics[width=1.\linewidth]{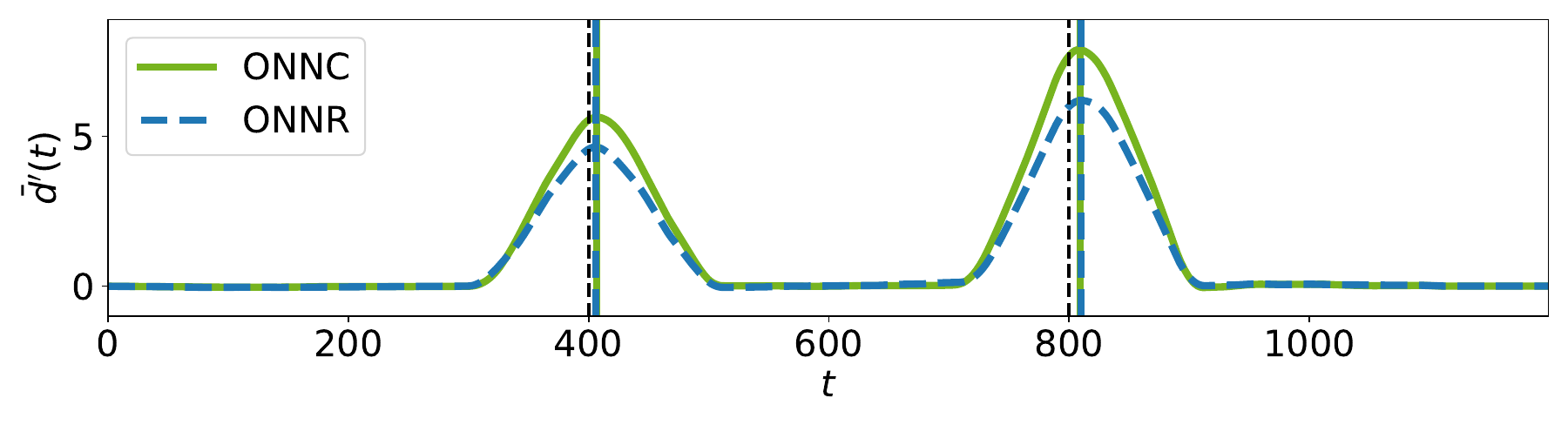}
\caption{Change-point detection score estimated by the algorithms ONNC and ONNR after the time shift: $\bar{d}'(t) = \bar{d}(t+l+n)$, where score $\bar{d}(t)$ is shown in Figure~\ref{fig:cpd_example_shift}. Positions of the score peaks are considered as positions of the detected change-points.}
\label{fig:cpd_example_shift}
\end{figure}

\begin{figure}[t]
\centering
\includegraphics[width=1.\linewidth]{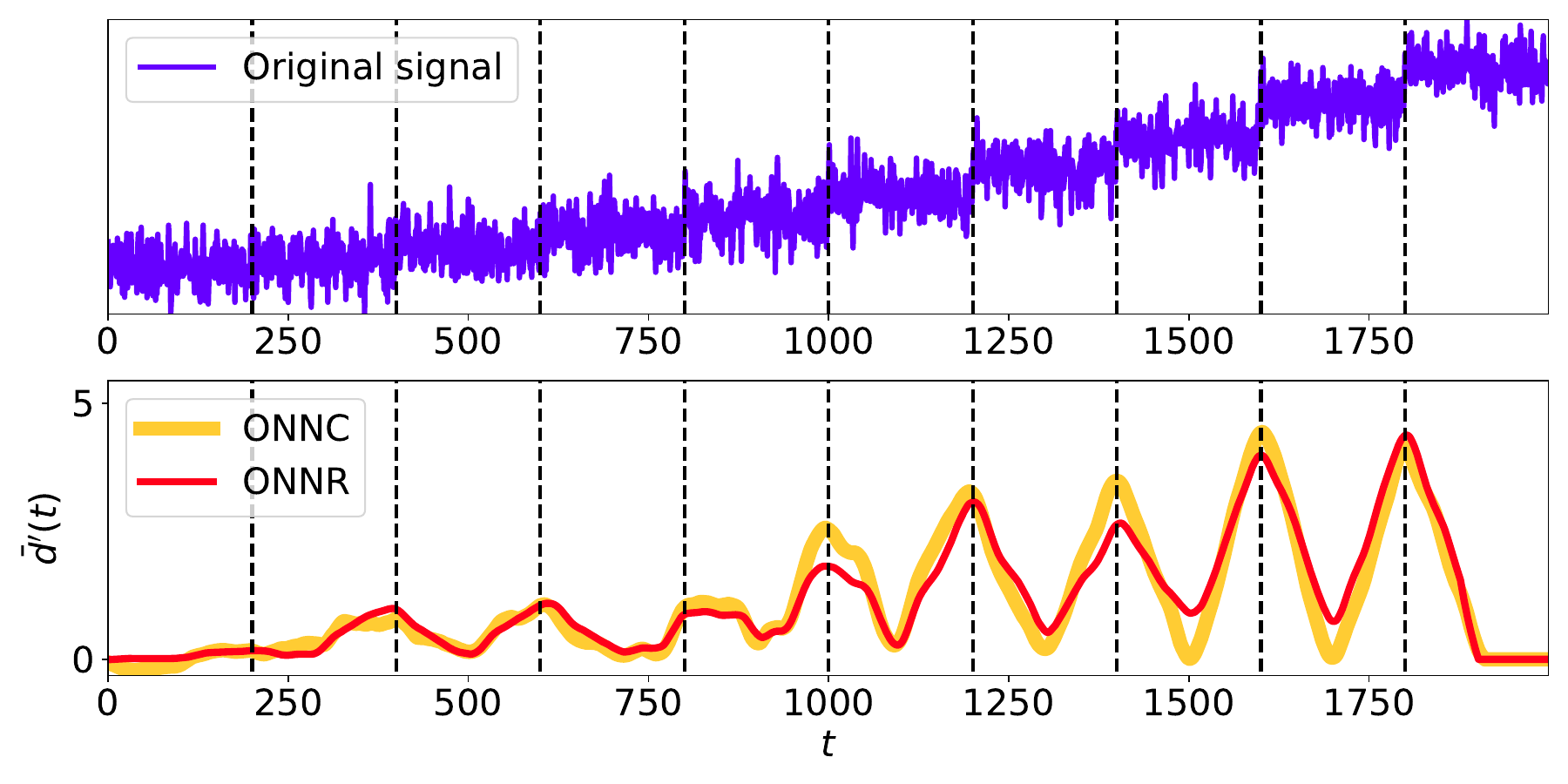}
\caption{Example of change-point detection score $\bar{d}'(t)$ estimated by ONNC and ONNR algorithms (bottom) for a time series in \textbf{mean jumps} data set (top).}
\label{fig:mean_jumps}
\end{figure}

We compare the proposed methods with 4 known methods for change-point detection\footnote{All code and data needed to reproduce our results are available in a repository: https://gitlab.com/lambda-hse/change-point/online-nn-cpd}. These methods are  Binseg~\cite{RePEc:cup:etheor:v:13:y:1997:i:03:p:315-352_00, fryzlewicz2014}, Pelt~\cite{Killick_2012}, Window~\cite{TRUONG2020107299} and RuLSIF~\cite{LIU201372}. There are several reviews~\cite{Aminikhanghahi2017, TRUONG2020107299, burg2020evaluation}, where it is shown that they demonstrate the best quality of change-point detection on various data sets. 

Implementations of Binseg, Pelt and Window algorithms in the ruptures~\cite{TRUONG2020107299} package are used in further experiments. The Binseg and Window methods require the set up of the number of change-points needed to be found in a time series. The optimal number for each sample is estimated from a range $[1, 40]$, using grid search, by maximizing RI quality metric. The Window algorithm also has \textit{width} hyperparameter. To provide good resolution between consecutive change points, we take $\textit{width}=20$ for Kepler, $\textit{width}=200$ for Higgs and $\textit{width}=100$ for the rest of the data sets described in Section~\ref{datasets}. Similarly, the Pelt method has a hyperparameter \textit{pen} for penalty. Its optimal value is found in the range $[0, 10]$ using grid search with step $0.5$ by maximizing the RI quality metric. For all these algorithms, we use the \textit{rbf} cost function as the most universal choice which works with any kind of change-points.

The regularisation parameter, $\lambda$, and width $\sigma$ of RBF kernels in the RuLSIF algorithm are also optimised using grid search in the range $[10^{-3}, 10^3]$. For the \textit{window size} hyperparameter, we take the same values as for the $\textit{width}$ hyperparameter in the Window algorithm.

For the proposed algorithms in this work, ONNC and ONNR, we use the following hyperparameters. The lag size $l=20$ for Kepler, $l=200$ for Higgs and $l=100$ for the rest of the data sets. The number of previous observations in~\eqref{eq:ar} $k=1$. The mini-batch size $n=\{1, 10\}$; the number of epochs of the neural network optimizer $n\_epochs = \{1, 10\}$ and the learning rate $lr = \{0.1, 0.01\}$. The optimal values of these hyperparameters are estimated using grid search by maximizing the RI quality metric. The neural network optimizer is Adam.

\begin{figure}[t]
\centering
\includegraphics[width=1.\linewidth]{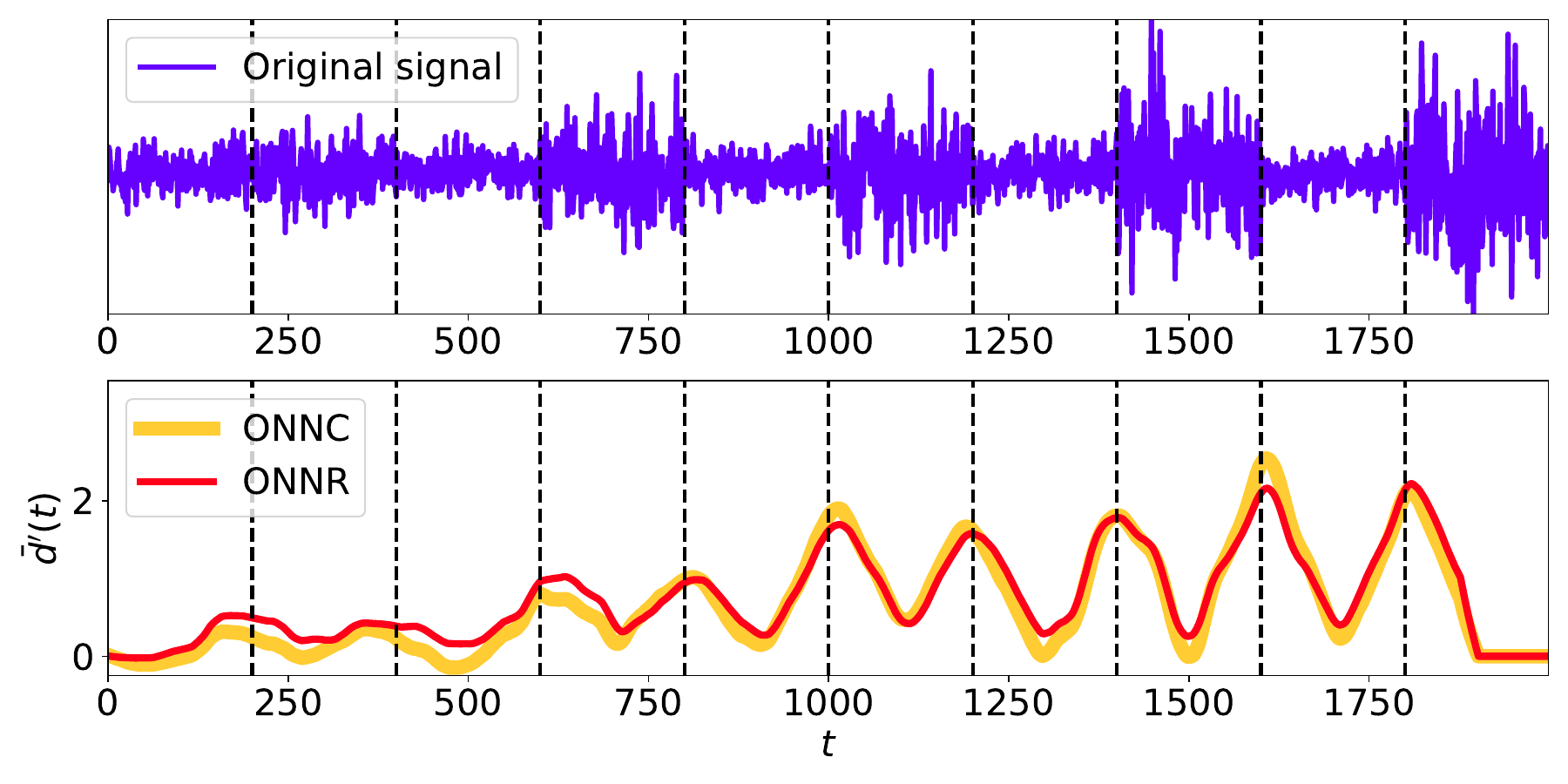}
\caption{Example of change-point detection score $\bar{d}'(t)$ estimated by ONNC and ONNR algorithms (bottom) for a time series in \textbf{variance jumps} data set (top).}
\label{fig:var_jumps}
\end{figure}

\begin{table}[!t]
\centering
\caption{Average values of RI quality metric for all change-point detection algorithms and data sets.}
\begin{tabular}{|l|rrrrrr|}
\hline
Dataset     &  Binseg &  Pelt &  Window &  RuLSIF &  ONNC &  ONNR \\
\hline
Mean jumps &      \textbf{0.99} &    \textbf{0.99} &      0.98 &         0.98 &            0.98 &      \textbf{0.99} \\
 Variance jumps &      \textbf{0.99} &    \textbf{0.99} &      0.98 &         0.98 &            0.98 &      0.98 \\
 Cov jumps &      \textbf{0.97} &    0.96 &      \textbf{0.97} &         0.95 &            \textbf{0.97} &      \textbf{0.97} \\
     MNIST &      \textbf{0.99} &    0.97 &      0.97 &         0.91 &            0.98 &      0.97 \\
     WISDM &      \textbf{0.99} &    \textbf{0.99} &      \textbf{0.99} &         \textbf{0.99} &            \textbf{0.99} &      \textbf{0.99} \\
       EMG &      0.97 &    0.97 &      0.97 &         0.97 &            \textbf{0.98} &      \textbf{0.98} \\
    Kepler &      0.95 &    0.99 &      0.99 &         0.89 &            \textbf{1.00} &      \textbf{1.00} \\
      SUSY &      \textbf{0.98} &    \textbf{0.98} &      0.97 &         0.95 &            \textbf{0.98} &      \textbf{0.98} \\
     Higgs &      0.96 &    0.91 &      0.95 &         0.75 &            \textbf{0.97} &      \textbf{0.97} \\
     MAGIC &      0.96 &    \textbf{0.97} &      \textbf{0.97} &         0.85 &            0.96 &      \textbf{0.97} \\
     HTRU2 &      \textbf{0.98} &    \textbf{0.98} &      0.97 &         0.96 &            \textbf{0.98} &      0.97 \\
\hline
\end{tabular}
\label{table:ri}
\end{table}

\begin{figure}[t]
\centering
\includegraphics[width=1.\linewidth]{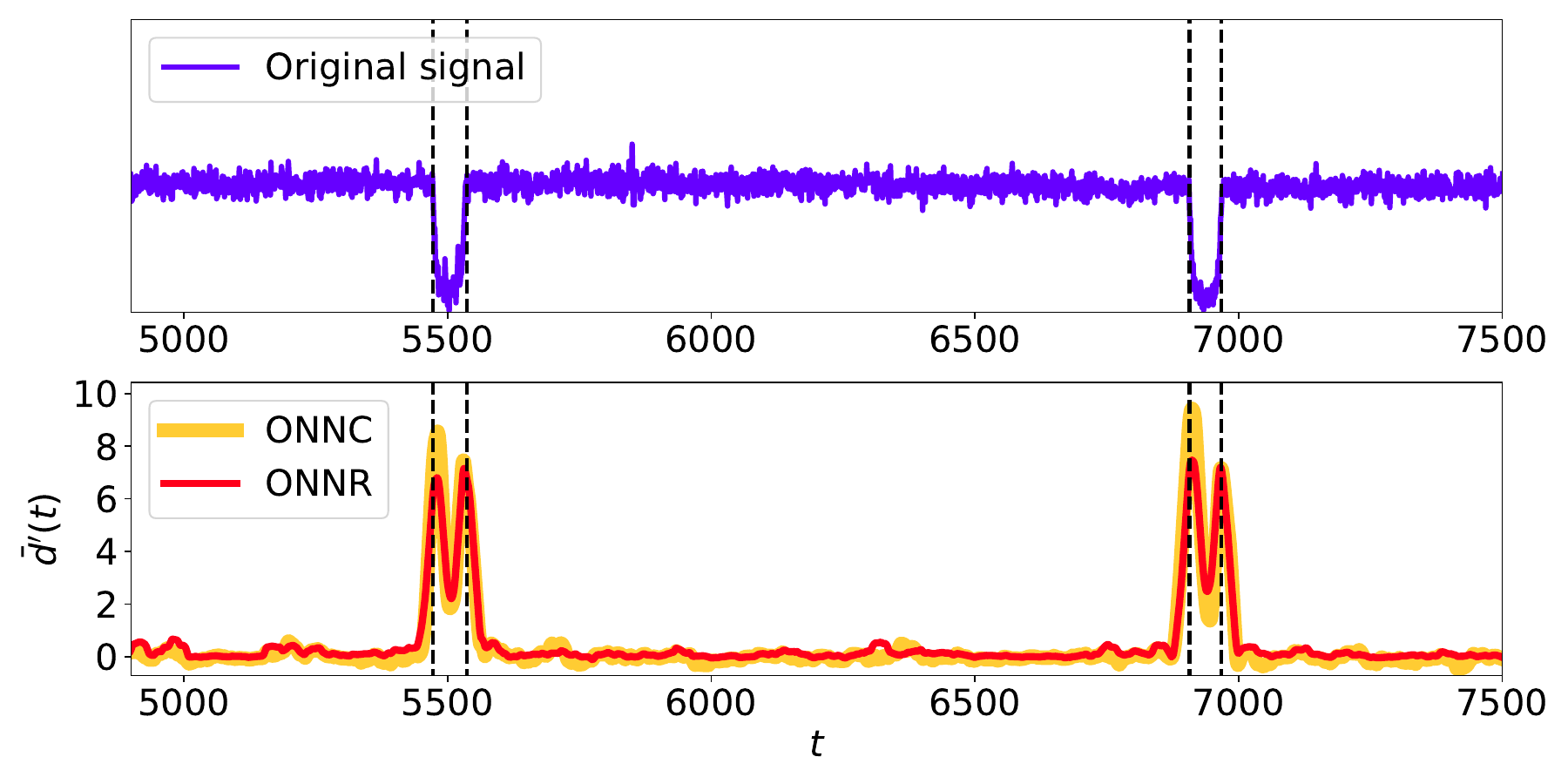}
\caption{Example of change-point detection score $\bar{d}'(t)$ estimated by ONNC and ONNR algorithms (bottom) for a time series in \textbf{Kepler} data set (top).}
\label{fig:kepler}
\end{figure}

Binseg, Pelt, Window and RuLSIF are offline algorithms for change-point detection. This means that they process observations of a time series in any order they need. It helps to detect change-points without time delay. Our algorithms are online and process the observations sequentially in time order. This creates a time delay in the change-point detection score $\bar{d}(t)$ as it is demonstrated in Figure~\ref{fig:cpd_example}. Assuming, that firstly, the whole time series is processed and then the quality is measured, we transform the score $\bar{d}(t)$ to the offline-equivalent form by applying time shift on the sum of the lag $l$ and mini-batch $n$ sizes: $\bar{d}'(t) = \bar{d}(t+l+n)$ as is shown in Figure~\ref{fig:cpd_example_shift}. Positions of the score peaks are considered as positions of the detected change-points.

\begin{table}[!t]
\centering
\caption{Average values of F1-score quality metric for all change-point detection algorithms and data sets.}
\begin{tabular}{|l|rrrrrr|}
\hline
Dataset     &  Binseg &  Pelt &  Window &  RuLSIF &  ONNC &  ONNR \\
\hline
Mean jumps &      0.94 &    0.92 &      0.93 &         \textbf{0.97} &            \textbf{0.97} &      \textbf{0.97} \\
 Variance jumps &      0.92 &    0.95 &      0.90 &         \textbf{0.97} &            \textbf{0.97} &      0.96 \\
 Cov jumps &      0.65 &    0.62 &      0.82 &         0.85 &            0.90 &      \textbf{0.93} \\
     MNIST &      \textbf{0.97} &    0.92 &      0.89 &         0.79 &            0.96 &      \textbf{0.97} \\
     WISDM &      0.88 &    0.86 &      0.94 &         0.94 &            0.96 &      \textbf{0.97} \\
       EMG &      0.90 &    0.89 &      0.82 &         0.95 &            \textbf{0.97} &      \textbf{0.97} \\
    Kepler &      0.60 &    0.97 &      0.88 &         0.14 &            \textbf{1.00} &      0.97 \\
      SUSY &      0.90 &    0.92 &      0.83 &         0.76 &            \textbf{0.99} &      0.97 \\
     Higgs &      0.51 &    0.18 &      0.52 &         0.23 &            \textbf{0.76} &      \textbf{0.76} \\
     MAGIC &      0.68 &    0.83 &      0.77 &         0.58 &            \textbf{0.88} &      0.87 \\
     HTRU2 &      0.91 &    0.90 &      0.82 &         0.85 &            \textbf{0.98} &      0.93 \\
\hline
\end{tabular}
\label{table:f1}
\end{table}

\begin{figure}[t]
\centering
\includegraphics[width=1.\linewidth]{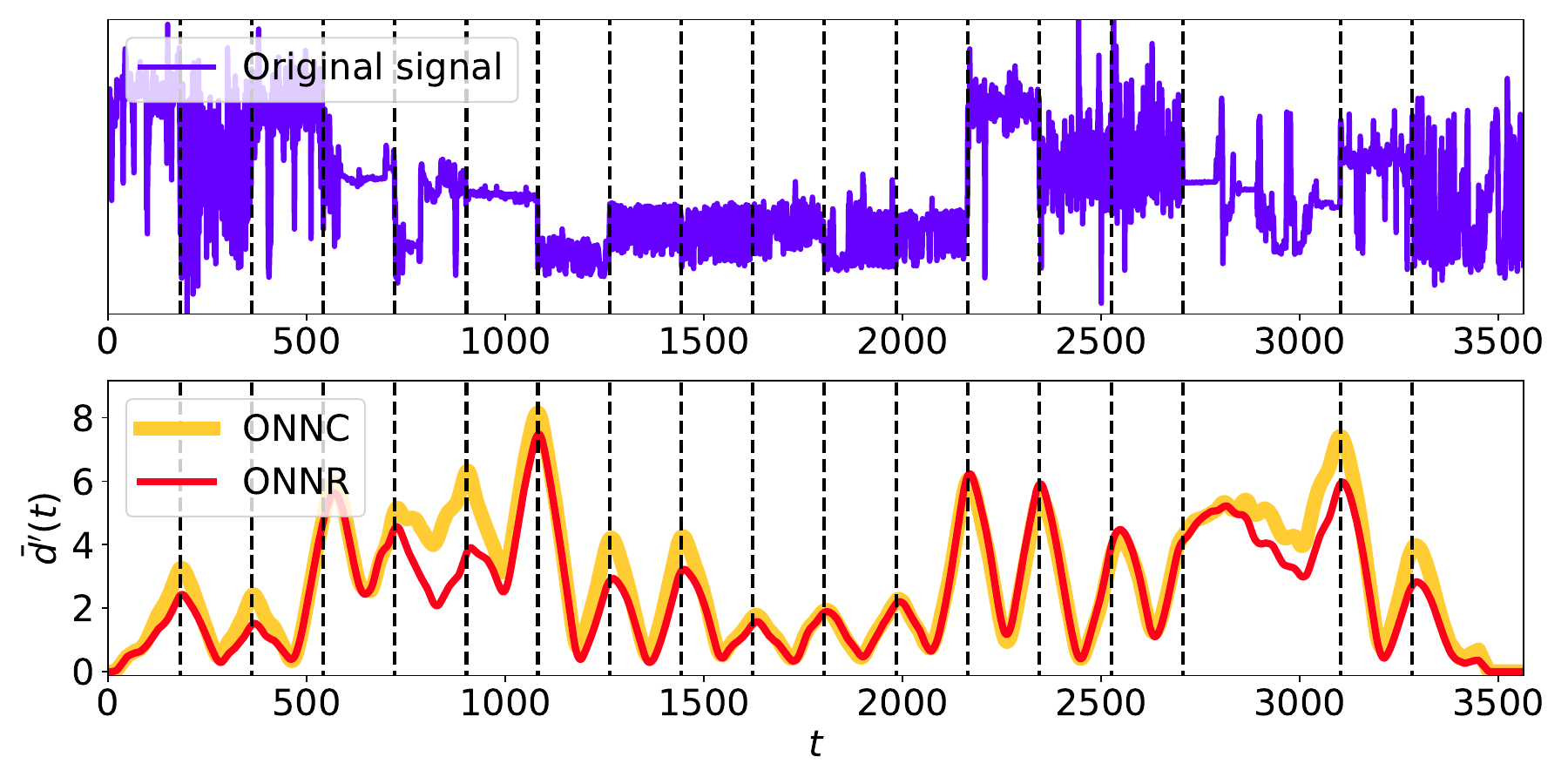}
\caption{Example of change-point detection score $\bar{d}'(t)$ estimated by ONNC and ONNR algorithms (bottom) for a time series in \textbf{WISDM} data set (top).}
\label{fig:wisdm}
\end{figure}

\begin{figure}[t]
\centering
\includegraphics[width=1.\linewidth]{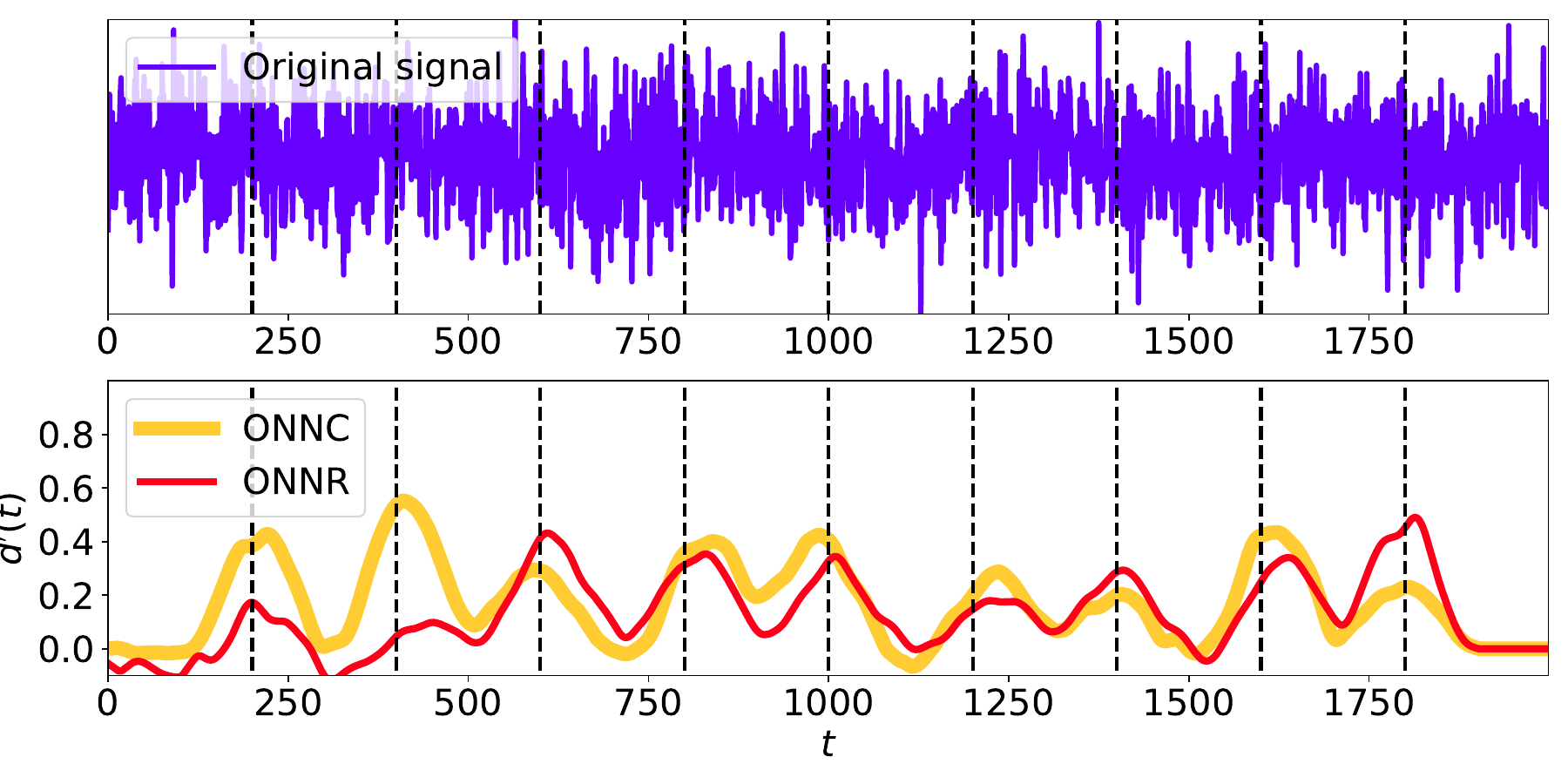}
\caption{Example of change-point detection score $\bar{d}'(t)$ estimated by ONNC and ONNR algorithms (bottom) for a time series in \textbf{HTRU2} data set (top).}
\label{fig:htru2}
\end{figure}

Each algorithm is applied to all time series in a data set. Then, the quality metric values are averaged over all samples in it. The average values of the RI and F1-score quality metrics are presented in Table~\ref{table:ri} and Table~\ref{table:f1} respectively. The results show that ONNC and ONNR have similar or better RI values for all data sets and demonstrate the best values of the F1-score for all data sets, except mean jumps and MNIST, where these algorithms show the same quality as other methods. Examples of change-point detection score, estimated by ONNC and ONNR algorithms, for several time series are demonstrated in Figure~\ref{fig:mean_jumps},~\ref{fig:var_jumps},~\ref{fig:kepler},~\ref{fig:wisdm} and~\ref{fig:htru2}.

\section{Discussion}

In this work, two new online algorithms for change-point detection in time series data are introduced. They are based on sequential comparison of two mini-batches of observations using neural networks, to estimate whether they have the same distribution. Each pair of mini-batches is processed only once, which provides good scalability of the algorithms.

The results in Table~\ref{table:ri} and Table~\ref{table:f1} demonstrate that the algorithms are able to detect various kinds of change-points in high-dimensional time series. Also, ONNC and ONNR methods demonstrate better quality of the detection on noisy data sets than other approaches. Reducing the noise level increases the quality for all algorithms considered here. To explain this, one can consider an RBF kernel for two observations $X(i)$ and $X(j)$ from Eq.~\eqref{eq:ts2}:

\begin{equation}
\label{eq:kernel}
    K(X(i), X(j)) = \exp(- \frac{d_{ij}^{2}}{2\sigma^{2}})
\end{equation}
and
\begin{equation}
    d_{ij} = \sqrt{(X_1(i) - X_1(j)^2 + ... + (X_{kd}(i) - X_{kd}(j)^2},
\end{equation}
where $\sigma$ is the kernel width; $d_{ij}$ is the Euclidean distance between the observations. The kernels are used in the cost functions of Binseg, Pelt, Window and RulSIF methods. In these equations, all signal components are taken into account equally. Uninformative and noisy components increase the variance of the distances, which reduces the sensitivity of the cost functions and decreases the quality of change-point detection.

\begin{table}[!t]
\centering
\caption{Computational complexity and memory usage of the change-point detection algorithms. $T$ - the number of observation in a time series; $W$ is the window width; $K$ is the number of kernels; $l$ is the lag size.}
\begin{tabular}{|l|r|r|}
\hline
{~~~~~~~~~~~~~~~}     &  ~~~~~~~~~~Computations &  ~~~~~~~~~~Memory \\
\hline
Binseg &      $\mathcal{O}(T^3)$ &    $\mathcal{O}(T^2)$ \\
  Pelt &      $\mathcal{O}(T^3)$ &    $\mathcal{O}(T^2)$  \\
Window &      $\mathcal{O}(W^2T)$ &    $\mathcal{O}(W^2)$  \\
RuLSIF &      $\mathcal{O}(KWT)$ &    $\mathcal{O}(KW)$  \\
  ONNC &      $\mathcal{O}(T)$ &    $\mathcal{O}(l)$  \\
  ONNR &      $\mathcal{O}(T)$ &    $\mathcal{O}(l)$  \\
\hline
\end{tabular}
\label{table:complexity}
\end{table}

As was considered previously, the ONNC and ONNR algorithms described in Algorithm~\ref{alg:clf} and Algorithm~\ref{alg:reg}, respectively, process mini-batches of a time series observations sequentially. Thus, the computational complexity of these methods is $\mathcal{O}(T)$, where $T$ is the total number of observations in the time series. They also need $\mathcal{O}(l)$ memory to store the last $l$ values of $d(t)$ score, where $l$ is the lag size between the mini-batches. This makes the ONNC and ONNR algorithms scalable and suitable for change-point detection in large time series.

According to~\cite{TRUONG2020107299}, the minimal theoretical computational complexities for Binseg and Pelt algorithms are $\mathcal{O}(T \log T)$ and $\mathcal{O}(T)$ respectively, for cases when the cost function requires $\mathcal{O}(1)$ operations on each step of the algorithms. However, using the cost function, based on RBF kernels, increases the required number of computations to $\mathcal{O}(T^3)$ and memory usage to $\mathcal{O}(T^2)$, due to the calculation of distances between pairs of observations. This makes them unsuitable for change-point detection in large time series. 

Similarly, the Window method needs $\mathcal{O}(W^2)$ operations at each step to calculate the pairwise distances between observations in windows with the width $W$. In the same way, RuLSIF requires $\mathcal{O}(KW)$ computations and memory at each step, where $K$ is the number of kernels used. Computational complexities and memory usage for the all algorithms considered in this paper are presented in Table~\ref{table:complexity}. It demonstrates that ONNC and ONNR algorithms are more scalable and take less computational resources than other methods.

\section{Conclusion}

In this work, we present two different change-point detection algorithms for time series data based on online learning. It is demonstrated that they outperform other popular algorithms on various synthetic and real-world data sets. The estimated computational complexities and memory usage show that they are faster than other methods, provide better scalability and are well suited for large time series for change-point detection. It is shown theoretically that the ONNC algorithm converges to its optimal solution. The discussion also defines the conditions, when the proposed online learning approach helps to achieve better solution than the offline one. We derived the exact equations of the optimal solutions for the both cases. They demonstrate the superiority of the online learning approach over the offline one for change point detection in time series.

Scripts of all our experiments in this work and all data sets are provided in the GitLab repository\footnote{\url{https://gitlab.com/lambda-hse/change-point/online-nn-cpd}}. The proposed change point detection algorithms using the online and offline learning are also implemented in Roerich python library\footnote{\url{https://github.com/HSE-LAMBDA/roerich}}.

\section{Acknowledgments}

The work was supported by the grant for research centers in the field of AI provided by the Ministry of Economic Development of the Russian Federation in accordance with the agreement 000000C313925P4E0002 and the agreement with HSE University № 139-15-2025-009.
The computation for this research was performed using the computational resources of HPC facilities at HSE University \cite{Kostenetskiy_2021}.

\begin{appendices}

\section{Optimal Predictions}

Suppose that $x(i) \sim p(x)$ for $t-N < i \le t$, and $x(i) \sim q(x)$ for $t-N-l < i \le t-l$, where $p(x)$ and $q(x)$ are some probability density functions. Define a discrete form $I_{N}$ of a binary cross-entropy loss function between two probability distributions $p(x)$ and $q(x)$ for a model $f(x, \theta)$, with weights $\theta$, and the continuous form $I$ of the function we define as

\begin{equation}
\label{eq:in}
\begin{split}
    I_{N} & = -\frac{1}{N} \sum_{i=t-N+1}^{t} L_{i}(\theta)
\end{split}
\end{equation}
\begin{equation}
\label{eq:i}
\begin{split}
    I & = \int p(x) \log f(x, \theta)  dx + \int q(x) \log (1-f(x, \theta)) dx
\end{split}
\end{equation}

The central limit theorem states that $I_N$ behaves as $I$ with quite large $N$ with the following variance convergence \cite{caflisch_1998}:
\begin{equation}
\label{eq:var}
\begin{split}
    Var_x[I_{N}] & = \mathbb{E}_x[(I_N - I)^2] = O(\frac{1}{N})
\end{split}
\end{equation}

During the model fitting we minimize the loss function value. Supposing that $\theta^{*} = \arg \max_{\theta} I$ and following the math in \cite{goodfellow2014generative} we obtain the optimal model predictions:
\begin{equation}
\label{eq:opt}
\begin{split}
    f(x, \theta^{*}) = \frac{p(x)}{p(x)+q(x)}
\end{split}
\end{equation}

\section{Proofs of Theorems}
\label{sec:proofs}

\begin{proof}[\textbf{The proof of Theorem \ref{lem:1}}]
Consider the following optimal value for $\theta$:
\begin{equation}
\theta_0 = \arg \min_{\theta} \sum_{i=t-N+1}^{t} L_{i}(\theta)
\end{equation}
Then
\begin{equation} 
\begin{split}
    \min_{\theta} \sum_{i=t-N+1}^{t} L_{i}(\theta) - \min_{\theta} \sum_{i=t-N+1}^{\nu} L_{i}(\theta) - \min_{\theta} \sum_{i=\nu+1}^{t} L_{i}(\theta) = \\
    (\sum_{i=t-N+1}^{\nu} L_{i}(\theta_0) - \min_{\theta} \sum_{i=t-N+1}^{\nu} L_{i}(\theta)) + (\sum_{i=\nu+1}^{t} L_{i}(\theta_0) - \min_{\theta} \sum_{i=\nu+1}^{t} L_{i}(\theta)) \ge 0 
\end{split}
\end{equation}
We use above the following property of minimum:
\begin{equation}
\sum_{i} L_{i}(\theta) - \min_{\theta} \sum_{i} L_{i}(\theta) \ge 0, \forall \theta
\end{equation}
\end{proof}

\begin{proof}[\textbf{The proof of Theorem \ref{theor:1}}]
According to~\cite{10.5555/3041838.3041955} we get the following upper bound for regret of OSD algorithm:
\begin{equation}
R(N) \le \frac{\parallel F \parallel^2}{2\eta} + \frac{\parallel \nabla L \parallel^2}{2} \eta N
\end{equation}
Similarly
\begin{equation}
\begin{split}
    R_1 & = \sum_{i=\nu+1}^{t} L_{i}(\theta_i) - \min_{\theta} \sum_{i=\nu+1}^{t} L_{i}(\theta) \\
        & \le \frac{\parallel F \parallel^2}{2\eta} + \frac{\parallel \nabla L \parallel^2}{2} \eta (t-\nu)
\end{split}
\end{equation}
\begin{equation}
\begin{split}
    R_2 & = \sum_{i=t-N+1}^{\nu} L_{i}(\theta_i) - \min_{\theta} \sum_{i=t-N+1}^{\nu} L_{i}(\theta) \\
        & \le \frac{\parallel F \parallel^2}{2\eta} + \frac{\parallel \nabla L \parallel^2}{2} \eta (\nu-t+N)
\end{split}
\end{equation}
Then
\begin{equation} 
\begin{split}
R(N) & = R_1 + R_2 - C(N, \nu) \\
     & \le \frac{\parallel F \parallel^2}{\eta} + \frac{\parallel \nabla L \parallel^2}{2} \eta N  - C(N, \nu)
\end{split}
\end{equation}
In result
\begin{equation} 
R(N) \le \frac{\parallel F \parallel^2}{\eta} + \frac{\parallel \nabla L \parallel^2}{2} \eta N - C(N, \nu)
\end{equation}
Theorem~\ref{lem:1} shows that $C(N, \nu) \ge 0$.

\end{proof}

\begin{proof}[\textbf{The proof of Theorem \ref{theor:2}}]

Let's expand the definition of the dissimilarity score into two terms:
\begin{equation} 
\begin{split}
    \bar{d}(t) & =\frac{1}{N} \sum_{t-N+1}^{t-\nu} \left( \log \frac{1 - f(x(i-N), \theta)}{f(x(i-N), \theta)} + \log \frac{f(x(i), \theta)}{1 - f(x(i), \theta)} \right) + \\ & + \frac{1}{N} \sum_{t-\nu+1}^{t} \left( \log \frac{1 - f(x(i-N), \theta)}{f(x(i-N), \theta)} + \log \frac{f(x(i), \theta)}{1 - f(x(i), \theta)} \right).
\end{split}
\end{equation}

According to Equation \ref{eq:opt}, the optimal solution for $t-N < i \le t-\nu$ corresponds to $f(x, \theta)=0.5$. Similarly, the optimal predictions for $t-\nu < i \le t$:

\begin{equation}
f(x(i), \theta) = \frac{p_1(x(i))}{p_1(x(i)) + p_0(x(i))}.
\end{equation}

Then

\begin{equation} 
\begin{split}
    \bar{d}(t) = \frac{1}{N} \sum_{t-\nu+1}^{t} \left( \log \frac{p_{0}(x(i-N))}{p_{1}(x(i-N))} + \log \frac{p_{1}(x(i))}{p_{0}(x(i))} \right).
\end{split}
\end{equation}

Taking the expected value from the expression above, we get the final result:

\begin{equation}
\mathbb{E}[\bar{d}(t)] = \frac{\nu}{N} \left( \mathbb{E}_{x\sim p_1(x)} \left[ \log \frac{p_1(x)}{p_0(x)} \right] - \mathbb{E}_{x\sim p_0(x)} \left[ \log \frac{p_1(x)}{p_0(x)} \right] \right).
\end{equation}

\end{proof}

\begin{proof}[\textbf{The proof of Theorem \ref{theor:3}}]

Let's expand the definition of the dissimilarity score into two terms:
\begin{equation} 
\begin{split}
    \bar{d}(t) & =\frac{1}{N} \sum_{t-N+1}^{t-\nu} \left( \log \frac{1 - f(x(i-N), \theta)}{f(x(i-N), \theta)} + \log \frac{f(x(i), \theta)}{1 - f(x(i), \theta)} \right) + \\ & + \frac{1}{N} \sum_{t-\nu+1}^{t} \left( \log \frac{1 - f(x(i-N), \theta)}{f(x(i-N), \theta)} + \log \frac{f(x(i), \theta)}{1 - f(x(i), \theta)} \right).
\end{split}
\end{equation}

According to Equation \ref{eq:opt}, the optimal prediction of the model:
\begin{equation}
f(x(i), \theta) = \frac{\tilde{p}_1(x(i))}{\tilde{p}_1(x(i)) + \tilde{p}_0(x(i))},
\end{equation}

where $\tilde{p}_0(x(i)) = p_0(x(i))$ and 
\begin{equation}
\tilde{p}_1(x(i)) = \frac{\nu p_1(x(i)) + (N-\nu) p_0(x(i))}{N}
\end{equation}

Then

\begin{equation} 
\begin{split}
    \bar{d}(t) = \frac{1}{N} \sum_{t-\nu+1}^{t} \left( \log \frac{\tilde{p}_{0}(x(i-N))}{\tilde{p}_{1}(x(i-N))} + \log \frac{\tilde{p}_{1}(x(i))}{\tilde{p}_{0}(x(i))} \right).
\end{split}
\end{equation}

Taking the expected value from the expression above, we get the final result:

\begin{equation}
\mathbb{E}[\bar{d}(t)] = \frac{\nu}{N} \left( \mathbb{E}_{x\sim p_1(x)} \left[ \log \frac{\tilde{p}_1(x)}{\tilde{p}_0(x)} \right] - \mathbb{E}_{x\sim p_0(x)} \left[ \log \frac{\tilde{p}_1(x)}{\tilde{p}_0(x)} \right] \right).
\end{equation}

\end{proof}




\end{appendices}


\bibliography{sn-bibliography}

@article{10.2307/2333009,
 ISSN = {00063444},
 URL = {http://www.jstor.org/stable/2333009},
 author = {E. S. Page},
 journal = {Biometrika},
 number = {1/2},
 pages = {100--115},
 publisher = {[Oxford University Press, Biometrika Trust]},
 title = {Continuous Inspection Schemes},
 volume = {41},
 year = {1954}
}

@article{10.1093/biomet/42.3-4.523,
    author = {PAGE, E. S.},
    title = "{A test for a change in a parameter occurring at an unknown point}",
    journal = {Biometrika},
    volume = {42},
    number = {3-4},
    pages = {523-527},
    year = {1955},
    month = {12},
    issn = {0006-3444},
    doi = {10.1093/biomet/42.3-4.523},
    url = {https://doi.org/10.1093/biomet/42.3-4.523},
    eprint = {https://academic.oup.com/biomet/article-pdf/42/3-4/523/838813/42-3-4-523.pdf},
}

@article{TRUONG2020107299,
title = "Selective review of offline change point detection methods",
journal = "Signal Processing",
volume = "167",
pages = "107299",
year = "2020",
issn = "0165-1684",
doi = "https://doi.org/10.1016/j.sigpro.2019.107299",
url = "http://www.sciencedirect.com/science/article/pii/S0165168419303494",
author = "Charles Truong and Laurent Oudre and Nicolas Vayatis"
}

@book{10.5555/151741,
author = {Basseville, Mich\`{e}le and Nikiforov, Igor V.},
title = {Detection of Abrupt Changes: Theory and Application},
year = {1993},
isbn = {0131267809},
publisher = {Prentice-Hall, Inc.},
address = {USA}
}

@inproceedings{kliep1,
author = {Sugiyama, Masashi and Nakajima, Shinichi and Kashima, Hisashi and von Bünau, Paul and Kawanabe, Motoaki},
year = {2007},
month = {01},
pages = {},
title = {Direct Importance Estimation with Model Selection and Its Application to Covariate Shift Adaptation.},
volume = {20},
journal = {Neural Info. Proc. Syst}
}

@article{ulsif1,
author = {Kanamori, Takafumi and Hido, Shohei and Sugiyama, Masashi},
year = {2009},
month = {07},
pages = {1391-1445},
title = {A Least-squares Approach to Direct Importance Estimation},
volume = {10},
journal = {Journal of Machine Learning Research},
doi = {10.1145/1577069.1755831}
}

@article{rulsif1,
author = {Kanamori, Takafumi and Suzuki, Taiji and Sugiyama, Masashi},
year = {2013},
month = {03},
pages = {},
title = {Computational Complexity of Kernel-Based Density-Ratio Estimation: A Condition Number Analysis},
volume = {90},
journal = {Machine Learning},
doi = {10.1007/s10994-012-5323-6}
}

@article{Sugiyama2011,
author = {Sugiyama, Masashi and Suzuki, Taiji and Kanamori, Takafumi},
year = {2011},
month = {10},
pages = {},
title = {Density Ratio Matching under the Bregman Divergence: A Unified Framework of Density Ratio Estimation},
volume = {64},
journal = {Annals of the Institute of Statistical Mathematics},
doi = {10.1007/s10463-011-0343-8}
}

@article{Yamada2013,
 author = {Yamada, Makoto and Suzuki, Taiji and Kanamori, Takafumi and Hachiya, Hirotaka and Sugiyama, Masashi},
 title = {Relative Density-ratio Estimation for Robust Distribution Comparison},
 journal = {Neural Comput.},
 issue_date = {May 2013},
 volume = {25},
 number = {5},
 month = may,
 year = {2013},
 issn = {0899-7667},
 pages = {1324--1370},
 numpages = {47},
 url = {https://doi.org/10.1162/NECO_a_00442},
 doi = {10.1162/NECO_a_00442},
 acmid = {3166185},
 publisher = {MIT Press},
 address = {Cambridge, MA, USA},
}

@article{LIU201372,
title = "Change-point detection in time-series data by relative density-ratio estimation",
journal = "Neural Networks",
volume = "43",
pages = "72 - 83",
year = "2013",
issn = "0893-6080",
doi = "https://doi.org/10.1016/j.neunet.2013.01.012",
url = "http://www.sciencedirect.com/science/article/pii/S0893608013000270",
author = "Song Liu and Makoto Yamada and Nigel Collier and Masashi Sugiyama"
}

@article{Nam2015,
author = {Nam, Hyunha and Sugiyama, Masashi},
year = {2015},
month = {05},
pages = {1073-1079},
title = {Direct Density Ratio Estimation with Convolutional Neural Networks with Application in Outlier Detection},
volume = {E98.D},
journal = {IEICE Transactions on Information and Systems},
doi = {10.1587/transinf.2014EDP7335}
}

@inproceedings{Hido2008,
 author = {Hido, Shohei and Id{\'e}, Tsuyoshi and Kashima, Hisashi and Kubo, Harunobu and Matsuzawa, Hirofumi},
 title = {Unsupervised Change Analysis Using Supervised Learning},
 booktitle = {Proceedings of the 12th Pacific-Asia Conference on Advances in Knowledge Discovery and Data Mining},
 series = {PAKDD'08},
 year = {2008},
 isbn = {3-540-68124-8, 978-3-540-68124-3},
 location = {Osaka, Japan},
 pages = {148--159},
 numpages = {12},
 url = {http://dl.acm.org/citation.cfm?id=1786574.1786592},
 acmid = {1786592},
 publisher = {Springer-Verlag},
 address = {Berlin, Heidelberg},
}

@misc{hushchyn2020generalization,
    title={Generalization of Change-Point Detection in Time Series Data Based on Direct Density Ratio Estimation},
    author={Mikhail Hushchyn and Andrey Ustyuzhanin},
    year={2020},
    eprint={2001.06386},
    archivePrefix={arXiv},
    primaryClass={cs.LG}
}

@misc{khan2019deep,
    title={Deep density ratio estimation for change point detection},
    author={Haidar Khan and Lara Marcuse and Bülent Yener},
    year={2019},
    eprint={1905.09876},
    archivePrefix={arXiv},
    primaryClass={cs.LG}
}

@Article{Aminikhanghahi2017,
author="Aminikhanghahi, Samaneh
and Cook, Diane J.",
title="A survey of methods for time series change point detection",
journal="Knowledge and Information Systems",
year="2017",
month="May",
day="01",
volume="51",
number="2",
pages="339--367",
issn="0219-3116",
doi="10.1007/s10115-016-0987-z",
url="https://doi.org/10.1007/s10115-016-0987-z"
}

@ARTICLE{8835065,
  author={G. M. {Weiss} and K. {Yoneda} and T. {Hayajneh}},
  journal={IEEE Access}, 
  title={Smartphone and Smartwatch-Based Biometrics Using Activities of Daily Living}, 
  year={2019},
  volume={7},
  number={},
  pages={133190-133202},}

@misc{Dua:2019 ,
author = "Dua, Dheeru and Graff, Casey",
year = "2017",
title = "{UCI} Machine Learning Repository",
url = "http://archive.ics.uci.edu/ml",
institution = "University of California, Irvine, School of Information and Computer Sciences" }

@article{10.1093/mnras/stw656,
    author = {Lyon, R. J. and Stappers, B. W. and Cooper, S. and Brooke, J. M. and Knowles, J. D.},
    title = "{Fifty years of pulsar candidate selection: from simple filters to a new principled real-time classification approach}",
    journal = {Monthly Notices of the Royal Astronomical Society},
    volume = {459},
    number = {1},
    pages = {1104-1123},
    year = {2016},
    month = {04},
    issn = {0035-8711},
    doi = {10.1093/mnras/stw656},
    url = {https://doi.org/10.1093/mnras/stw656},
    eprint = {https://academic.oup.com/mnras/article-pdf/459/1/1104/8115310/stw656.pdf},
}

@article{doi:10.1111/j.1365-2966.2010.17325.x,
author = {Keith, M. J. and Jameson, A. and van Straten, W. and Bailes, M. and Johnston, S. and Kramer, M. and Possenti, A. and Bates, S. D. and Bhat, N. D. R. and Burgay, M. and Burke-Spolaor, S. and D'Amico, N. and Levin, L. and McMahon, Peter L. and Milia, S. and Stappers, B. W.},
title = {The High Time Resolution Universe Pulsar Survey – I. System configuration and initial discoveries},
journal = {Monthly Notices of the Royal Astronomical Society},
volume = {409},
number = {2},
pages = {619-627},
doi = {10.1111/j.1365-2966.2010.17325.x},
url = {https://onlinelibrary.wiley.com/doi/abs/10.1111/j.1365-2966.2010.17325.x},
eprint = {https://onlinelibrary.wiley.com/doi/pdf/10.1111/j.1365-2966.2010.17325.x},
year = {2010}
}

@article{Baldi_2014,
   title={Searching for exotic particles in high-energy physics with deep learning},
   volume={5},
   ISSN={2041-1723},
   url={http://dx.doi.org/10.1038/ncomms5308},
   DOI={10.1038/ncomms5308},
   number={1},
   journal={Nature Communications},
   publisher={Springer Science and Business Media LLC},
   author={Baldi, P. and Sadowski, P. and Whiteson, D.},
   year={2014},
   month={Jul}
}

@misc{kepler2019,
  author = {\text{Kepler and K2 Science Center}},
  title = {\text{Kepler and K2 data products}},
  year = {2019},
  note = {\url{https://keplerscience.arc.nasa.gov/data-products.html}},
}

@ARTICLE{RePEc:cup:etheor:v:13:y:1997:i:03:p:315-352_00,
title = {Estimating Multiple Breaks One at a Time},
author = {Bai, Jushan},
year = {1997},
journal = {Econometric Theory},
volume = {13},
number = {3},
pages = {315-352},
url = {https://EconPapers.repec.org/RePEc:cup:etheor:v:13:y:1997:i:03:p:315-352_00}
}

@article{fryzlewicz2014,
author = "Fryzlewicz, Piotr",
doi = "10.1214/14-AOS1245",
fjournal = "Annals of Statistics",
journal = "Ann. Statist.",
month = "12",
number = "6",
pages = "2243--2281",
publisher = "The Institute of Mathematical Statistics",
title = "Wild binary segmentation for multiple change-point detection",
url = "https://doi.org/10.1214/14-AOS1245",
volume = "42",
year = "2014"
}

@article{Killick_2012,
   title={Optimal Detection of Changepoints With a Linear Computational Cost},
   volume={107},
   ISSN={1537-274X},
   url={http://dx.doi.org/10.1080/01621459.2012.737745},
   DOI={10.1080/01621459.2012.737745},
   number={500},
   journal={Journal of the American Statistical Association},
   publisher={Informa UK Limited},
   author={Killick, R. and Fearnhead, P. and Eckley, I. A.},
   year={2012},
   month={Oct},
   pages={1590–1598}
}

@misc{burg2020evaluation,
    title={An Evaluation of Change Point Detection Algorithms},
    author={Gerrit J. J. van den Burg and Christopher K. I. Williams},
    year={2020},
    eprint={2003.06222},
    archivePrefix={arXiv},
    primaryClass={stat.ML}
}

@misc{goodfellow2014generative,
      title={Generative Adversarial Networks}, 
      author={Ian J. Goodfellow and Jean Pouget-Abadie and Mehdi Mirza and Bing Xu and David Warde-Farley and Sherjil Ozair and Aaron Courville and Yoshua Bengio},
      year={2014},
      eprint={1406.2661},
      archivePrefix={arXiv},
      primaryClass={stat.ML}
}

@article{caflisch_1998, title={Monte Carlo and quasi-Monte Carlo methods}, volume={7}, DOI={10.1017/S0962492900002804}, journal={Acta Numerica}, publisher={Cambridge University Press}, author={Caflisch, Russel E.}, year={1998}, pages={1–49}}

@inproceedings{10.5555/3041838.3041955,
author = {Zinkevich, Martin},
title = {Online Convex Programming and Generalized Infinitesimal Gradient Ascent},
year = {2003},
isbn = {1577351894},
publisher = {AAAI Press},
booktitle = {Proceedings of the Twentieth International Conference on International Conference on Machine Learning},
pages = {928–935},
numpages = {8},
location = {Washington, DC, USA},
series = {ICML'03}
}

@article{rand_index,
author = { William M.   Rand },
title = {Objective Criteria for the Evaluation of Clustering Methods},
journal = {Journal of the American Statistical Association},
volume = {66},
number = {336},
pages = {846-850},
year  = {1971},
publisher = {Taylor & Francis},
doi = {10.1080/01621459.1971.10482356},

URL = { 
        https://www.tandfonline.com/doi/abs/10.1080/01621459.1971.10482356
    
},
eprint = { 
        https://www.tandfonline.com/doi/pdf/10.1080/01621459.1971.10482356
    
}

}

@article{Kostenetskiy_2021,
doi = {10.1088/1742-6596/1740/1/012050},
url = {https://dx.doi.org/10.1088/1742-6596/1740/1/012050},
year = {2021},
month = {jan},
publisher = {IOP Publishing},
volume = {1740},
number = {1},
pages = {012050},
author = {P. S. Kostenetskiy and R. A. Chulkevich and V. I. Kozyrev},
title = {HPC Resources of the Higher School of Economics},
journal = {Journal of Physics: Conference Series}
}

\end{document}